\documentclass[letterpaper, 10 pt, conference]{ieeeconf}  % Comment this line out if you need a4paper

\IEEEoverridecommandlockouts                              % This command is only needed if 
\usepackage{graphicx}
\usepackage{float} % optional, for [H]
\usepackage{hyperref} % for \url and hyperlinks
\usepackage{amsmath}
\usepackage{booktabs}
\title{\LARGE \bf
A Pilot Benchmark for NL-to-FOL Translation in Planetary Exploration
}

\author{Hayden Moore$^{1}$, Suman Saha$^{1}$, and Mahfuza Farooque$^{1}$% <-this % stops a space
\thanks{$^{1}$Department of Computer Science and Engineering, College of Engineering,
The Pennsylvania State University, University Park, Pennsylvania 16802, USA: {\tt\small hmm5731@psu.edu}}%
}

\begin{document}

\maketitle
\thispagestyle{empty}
\pagestyle{empty}

%%%%%%%%%%%%%%%%%%%%%%%%%%%%%%%%%%%%%%%%%%%%%%%%%%%%%%%%%%%%%%%%%%%%%%%%%%%%%%%%
\begin{abstract}
Future planetary exploration envisions autonomous robotic agents operating under severe communication constraints, without global positioning, and with minimal human intervention. In such environments, agents must not only perceive and act, but also reason over mission objectives, operational constraints, and evolving environmental conditions. While prior work has largely focused on perception and control, the translation of high-level mission knowledge into structured, machine-interpretable representations remains underexplored.

We introduce a pilot benchmark for translating natural language (NL) into First-Order Logic (FOL) within the domain of planetary exploration. The dataset is constructed from real mission documentation sourced from NASA’s Planetary Data System (PDS), spanning missions from 2003 to 2013. These documents describe mission phases such as launch, boost, coast, cruise, and orbital operations in rich natural language. We manually annotate these documents with corresponding FOL representations that capture temporal structure, agent roles, and operational dependencies. In addition, we provide structured predicate vocabularies and typed constants to enable controlled experimentation with varying levels of prior knowledge. This pilot benchmark provides a foundation for research at the intersection of language understanding and formal reasoning, grounded in real-world, safety-critical mission data. The dataset is provided at: \url{https://github.com/HaydenMM/planetary-logic-benchmark/blob/main/pilot_benchmark.json}.
\end{abstract}

%%%%%%%%%%%%%%%%%%%%%%%%%%%%%%%%%%%%%%%%%%%%%%%%%%%%%%%%%%%%%%%%%%%%%%%%%%%%%%%%
\section{INTRODUCTION}
Autonomous robotic systems are expected to play a central role in the next generation of planetary exploration \cite{c1}. Unlike terrestrial systems, these agents must operate under extreme constraints, including delayed or intermittent communication with Earth \cite{c3}, limited onboard computational resources, and the absence of global positioning systems. In such conditions, autonomy is not merely a convenience but a necessity. Agents must interpret mission objectives, reason about evolving environmental conditions, and make decisions that are both safe and effective without continuous human oversight.

A key challenge in achieving this level of autonomy lies in how mission knowledge is represented and utilized. In current practice, mission plans, operational procedures, and system constraints are largely expressed in natural language. These descriptions contain rich information about temporal sequences, dependencies between subsystems, and conditional behaviors. However, natural language lacks the precision and structure required for reliable execution and formal reasoning. As a result, there is a gap between how mission knowledge is specified and how it can be consumed by autonomous systems.

\begin{figure}[t]
    \centering
    \includegraphics[width=0.9\columnwidth]{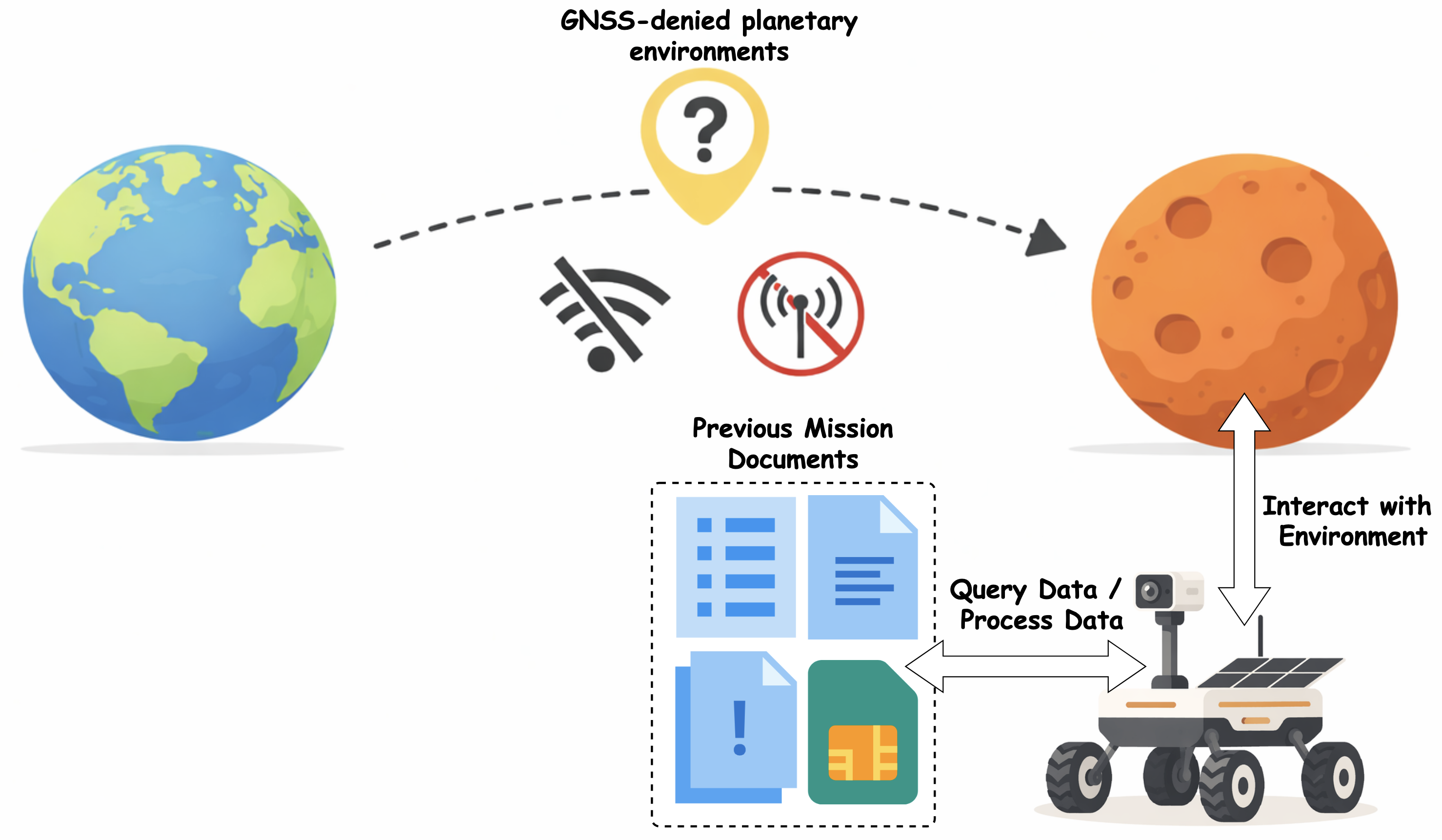}
    \caption{Conceptual setting motivating this work. In long-duration, communication-constrained planetary exploration, an autonomous agent cannot rely on continuous connectivity to Earth and must instead use onboard mission documents and internal reasoning to interpret mission phases, understand constraints, and make decisions while interacting with its environment.}
    \label{fig:intro_pipeline}
\end{figure}

Bridging this gap requires transforming high-level natural language descriptions into structured, machine-interpretable representations. First-Order Logic (FOL) provides a natural candidate for this transformation, offering a formal framework for representing entities, relations, and temporal dependencies. Despite its relevance, the problem of translating natural language mission descriptions into FOL has received limited attention, particularly in domains characterized by long-horizon temporal reasoning and complex operational structure \cite{c4,c5,c21,c22,c23}.
Figure~\ref{fig:intro_pipeline} illustrates the motivating setting for this benchmark, where an autonomous planetary agent must reason over onboard mission knowledge under communication constraints.

In this work, we introduce a pilot benchmark designed to support research on this problem. Our goal is to provide a dataset that captures the richness and complexity of real planetary mission documentation while enabling systematic exploration of natural language to logical representation. Rather than focusing on evaluating specific models, we position this benchmark as a foundational resource for studying the interaction between language understanding and formal reasoning in safety-critical planetary exploration environments. This complements prior work in autoformalization and structured translation tasks \cite{c21,c22,c23}.
\section{DATASET CONSTRUCTION}
The dataset is constructed from publicly available mission documentation provided by NASA’s Planetary Data System (PDS) \cite{c25}. These documents are written for scientific and operational purposes and contain detailed descriptions of mission phases, system behaviors, and observational activities. As such, they provide a realistic and challenging source of natural language data for structured reasoning tasks. The full dataset is available for anonymous review at: \url{https://anonymous.4open.science/r/PMR-BF96/mission.json}.

We collect mission documents spanning the period from 2003 to 2013, covering a diverse set of planetary missions and operational contexts. Each document describes the progression of a mission through multiple phases, including launch, boost, coast, cruise, orbit insertion, and science operations. These phases vary in duration from minutes to years and often involve nested temporal relationships and interdependent activities. The resulting dataset captures both short-duration procedural events and long-horizon mission planning, providing a broad range of reasoning challenges.

Each document is manually translated into a corresponding FOL representation. This annotation process is designed to preserve the semantic content of the original text while making its structure explicit. Temporal relationships are encoded using predicates that capture ordering, duration, and dependency, allowing events to be composed into coherent logical sequences. Agent roles are explicitly represented, distinguishing between spacecraft, subsystems, instruments, and external entities such as ground stations. Operational activities, such as maneuvers, observations, and system checks, are modeled as predicates that relate agents, actions, and outcomes.

An example from the dataset illustrates the level of detail captured in these annotations. The following excerpt shows a natural language description of a launch sequence alongside its corresponding First-Order Logic representation:

\noindent \textbf{Natural Language:} \\
\begin{quote}
Spirit (MER-2 hardware) was launched June 10, 2003, at 1759 UTC (1359 EDT) from launch complex 17A at Cape Canaveral Air Force Station, Florida. The launch azimuth was 93 degrees. The boost portion of the launch vehicle trajectory took approximately 10 minutes, and was followed by a short coast phase in a parking orbit for approximately 15 minutes. After third stage burnout, the upper stage despun the stack using a yo-yo despin system. Separation of the third stage occurred approximately 36 minutes after launch.
\end{quote}

\noindent \textbf{First-Order Logic:} \\
\begin{quote}
$\exists e \; ( \text{LaunchEvent}(e) \land \text{Agent}(e,\text{Spirit\_MER2}) \land \text{DateUTC}(e,'2003\text{-}06\text{-}10') \land \text{TimeUTC}(e,'17{:}59') \land \text{LocalTime}(e,'13{:}59') \land \text{LocalTZ}(e,'EDT') \land \text{From}(e,\text{LaunchComplex17A}) \land \text{At}(e,\text{CapeCanaveralAFS}) \land \text{In}(e,\text{Florida}) \land \text{AzimuthDeg}(e,93) )$ \\
$\land \exists b \; ( \text{BoostPhase}(b) \land \text{Agent}(b,\text{Spirit\_MER2}) \land \text{StartsAfterMinutesApprox}(b,e,0) \land \text{DurationMinutesApprox}(b,10) )$ \\
$\land \exists c \; ( \text{CoastPhase}(c) \land \text{Agent}(c,\text{Spirit\_MER2}) \land \text{InOrbit}(c,\text{ParkingOrbit}) \land \text{StartsAfterMinutesApprox}(c,b,0) \land \text{DurationMinutesApprox}(c,15) \land \text{After}(c,b) )$ \\
$\land \exists d \; ( \text{DespinEvent}(d) \land \text{Agent}(d,\text{Spirit\_MER2}) \land \text{After}(d,\text{ThirdStageBurnout}) \land \text{Uses}(d,\text{YoYoDespinSystem}) \land \text{ActsOn}(d,\text{UpperStageStack}) )$ \\
$\land \exists s \; ( \text{ThirdStageSeparation}(s) \land \text{Agent}(s,\text{Spirit\_MER2}) \land \text{AfterMinutesApprox}(s,e,36) )$
\end{quote}

Beyond the paired natural language and FOL representations, the dataset includes additional structure that supports controlled experimentation. Predicate vocabularies are extracted and standardized across all documents, providing a consistent set of logical primitives. Constants are separated and typed according to their semantic roles, including entities, temporal values, and numerical quantities. This separation allows the dataset to be used in settings with varying degrees of prior knowledge, such as scenarios where the logical vocabulary is fixed or where type information is available to guide parsing.

\section{BENCHMARK DESIGN}

The pilot benchmark is defined as the task of translating natural language mission descriptions into First-Order Logic representations. This task requires capturing not only the meaning of individual statements but also the compositional structure that arises from temporal dependencies and operational constraints. Unlike traditional semantic parsing tasks, which often focus on short and relatively independent utterances, this benchmark involves longer documents with interconnected events and hierarchical structure.

A central feature of the benchmark is its modular design. By explicitly separating predicate vocabularies and typed constants from the core annotations, the dataset enables researchers to explore different problem formulations. In one setting, models may be required to generate fully unconstrained logical forms directly from natural language. In another, models may operate within a predefined schema, leveraging known predicates and type constraints to guide the translation process. These variations allow for systematic investigation of how structure and prior knowledge influence performance.

The benchmark is intentionally agnostic to specific evaluation protocols. Instead, it is designed to support a range of research directions, including symbolic methods that rely on rule-based parsing, neural approaches that learn mappings from data, and hybrid methods that combine both paradigms \cite{c9,c10}. By focusing on the dataset itself, we aim to provide a flexible foundation that can be adapted to different experimental goals.

\section{MOTIVATION AND APPLICATION}
The motivation for this pilot benchmark is rooted in the operational requirements of future autonomous planetary systems. In scenarios where communication delays prevent real-time human intervention, agents must rely on internal representations of mission knowledge to make decisions. These representations must be both expressive enough to capture complex behaviors and structured enough to support reliable reasoning.

First-Order Logic offers a formalism that satisfies these requirements. In the context of planetary rover autonomy, this capability enables several concrete functions. Structured logical representations derived from natural language mission descriptions can be used to define constraints for semantic mapping, inform task planning and scheduling, enforce safety conditions during navigation, and support verification and validation (V\&V) of autonomous decisions. By grounding high-level mission intent in formal representations, autonomous systems can more reliably interpret objectives, reason over constraints, and ensure that executed actions remain consistent with mission requirements.

The benchmark also has implications beyond planetary exploration. Many domains require the translation of natural language into structured representations, including robotics, aerospace systems, and industrial automation. In these settings, the ability to bridge language and logic can enable more interpretable and reliable systems. By grounding this problem in real-world mission data, the benchmark provides a realistic testbed for developing and evaluating such capabilities \cite{c11,c12}.

\section{BENCHMARK STATISTICS}

The pilot benchmark currently consists of 11 multi-phase mission segments in natural language (NL) and their corresponding First-Order Logic (FOL). These examples are derived from 5 unique planetary mission sources spanning NASA Planetary Data System (PDS) documentation from 2003 to 2013. Across the dataset, we identify 251 unique predicate signatures and 125 distinct entity constants, along with 16 date literals, 13 time literals, and 51 numeric values. This reflects both the semantic diversity of mission descriptions and the structured variability required for formal reasoning.

The natural language inputs are moderately long and descriptive, with an average length of 257.6 words (median 249), ranging from 84 to 412 words. The corresponding FOL representations are similarly expressive, averaging 286.9 tokens (median 288), with lengths ranging from 174 to 434 tokens. This near parity in NL and FOL length highlights that the translation task preserves substantial structural detail rather than compressing information into shallow forms.

At the logical level, each example contains an average of 43.4 predicate mentions and 13.0 quantifiers, indicating a high degree of compositional structure. The number of unique predicates per example averages 30.3, suggesting that each mission segment introduces a rich and varied set of relations. Entity constants average 12.2 per example, while numeric constants average 4.6, reflecting the prevalence of temporal durations, orbital parameters, and measurement values in mission descriptions.

The dataset also exhibits significant variability in complexity. Natural language inputs range from short procedural descriptions (84 words) to long, multi-phase mission narratives (over 400 words). Correspondingly, FOL representations vary in both depth and breadth, with predicate vocabularies ranging from 17 to 50 per example and predicate mentions ranging from 27 to 65. Quantifier counts range from 5 to 25, further emphasizing differences in logical nesting and compositional depth across mission segments.

An analysis of predicate usage reveals consistent structural patterns across examples. The most frequent predicates include \texttt{During} (42 occurrences), \texttt{Agent} (26), \texttt{DateUTC} (13), and temporal relation predicates such as \texttt{StartsAt}, \texttt{EndsAt}, and \texttt{After}. These distributions highlight the central role of temporal reasoning and agent-centric event modeling in the dataset.

Overall, these statistics indicate that the benchmark captures both linguistic and logical complexity, combining long-form natural language descriptions with deeply structured formal representations. The diversity in predicate vocabularies, temporal constructs, and entity types makes this dataset well-suited for studying compositional generalization and structured reasoning in realistic, safety-critical and planetary exploration domains. Table~\ref{tab:benchmark_stats} summarizes the main dataset-level statistics, including natural language length, FOL length, predicate usage, quantifier count, and entity constants.

\begin{table}[h]
\centering
\small
\begin{tabular}{lcc}
\toprule
Statistic & Mean & Range \\
\midrule
NL Length (words) & 257.6 & 84–412 \\
FOL Length (tokens) & 286.9 & 174–434 \\
Predicate Mentions & 43.4 & 27–65 \\
Unique Predicates / Example & 30.3 & 17–50 \\
Quantifiers & 13.0 & 5–25 \\
Entity Constants & 12.2 & 6–18 \\
\bottomrule
\end{tabular}
\caption{Summary statistics of the NL–FOL pilot benchmark.}
\label{tab:benchmark_stats}
\end{table}

\section{PRELIMINARY RESULTS}

To provide an initial characterization of the pilot benchmark, we conducted exploratory experiments using a set of publicly available local language models, including \texttt{Qwen/Qwen1.5-1.8B-Chat} \cite{c13}, \texttt{mistralai/Mistral-7B-Instruct-v0.3} \cite{c14}, \texttt{microsoft/Phi-3-mini-4k-instruct} \cite{c15}, \texttt{meta-llama/Meta-Llama-3.1-8B-Instruct} \cite{c16}, and \texttt{google/gemma-2-9b-it} \cite{c17}. For each model, we provided the full natural language mission description as input and prompted the model to directly generate a corresponding First-Order Logic (FOL) representation without additional explanation or intermediate steps. All model outputs were fixed to a maximum of 500 tokens.

This setup represents a straightforward, unconstrained translation task and is not intended to reflect optimized prompting or system design. Rather, it serves as a baseline to illustrate the challenges posed by long-form, structured mission descriptions.

Across all models, we observe consistent failure modes. First, models struggle to maintain temporal and logical consistency over long sequences, often producing incomplete or incoherent event structures. Temporal relationships such as ordering, duration, and dependency are frequently omitted, misrepresented, or contradicted within the generated outputs. Second, models tend to miss critical components embedded within the natural language descriptions, particularly when these components appear in later portions of the text or are nested within complex sentences. Third, hallucination is prevalent, with models introducing predicates, entities, or relationships that are not grounded in the input description. These issues are amplified by the length and compositional structure of the task, which requires sustained reasoning across multiple interconnected events.

Some of the models demonstrate improved fluency and partial structure, but still fail to reliably capture the full set of temporal dependencies and operational constraints present in the source text. Notably, even when local consistency is maintained within short segments, global coherence across the full mission sequence remains a challenge.

Model-specific behaviors further highlight these limitations. Qwen \cite{c13} frequently fails to produce valid FOL representations and instead hallucinates by outputting natural language instructions or step-by-step procedural descriptions. Mistral \cite{c14} and Meta-LLaMA-3.1 \cite{c16} demonstrate stronger alignment with the task and are more likely to attempt FOL-like structures, but still tend to generate ordered lists of components that require additional processing to convert into valid logical forms. Interestingly, while these enumerated outputs are incorrect with respect to the target format, they often preserve temporal sequencing (e.g., step-wise progression), which partially maintains logical ordering despite not being explicitly requested.

Microsoft Phi-3 \cite{c15} exhibits severe hallucination behavior, frequently introducing unrelated space or NASA concepts not present in the input (e.g., references to the Hubble Space Telescope). In addition, it often repeats hallucinated temporal statements such as “Launch-40 min” and generates redundant natural language outputs describing tasks rather than producing FOL. These repetitions further degrade output quality and consistency.

In addition, Google Gemma \cite{c17} occasionally failed to produce any output when given the full mission descriptions, returning either empty responses or terminating generation prematurely. This behavior was observed more frequently for longer inputs and suggests limitations in handling extended context windows or internal decoding constraints under long-form generation.

Among the evaluated models, Meta-LLaMA-3.1 show the strongest ability to produce structured FOL-like outputs in some cases. However, even in these instances, the generated representations deviate significantly from the ground truth in both syntax and structure. 

These preliminary results highlight the difficulty of direct, single-pass translation from long-form natural language into structured logical representations. They suggest that more effective approaches will likely require decomposition strategies, such as breaking mission descriptions into smaller units, as well as structured prompting techniques. Potential directions include few-shot prompting with schema constraints, guided decoding with formal grammars, and hybrid neuro-symbolic pipelines that enforce logical consistency during generation \cite{c18,c19,c20}.

We emphasize that these experiments are not intended as a comprehensive evaluation, but rather as an initial demonstration of the challenges inherent in this pilot benchmark. All raw model outputs are available at: \url{https://github.com/HaydenMM/planetary-logic-benchmark/blob/main/pilot_benchmark_output.txt}.

\section{LIMITATIONS}

While this work introduces a curated pilot benchmark for NL-to-FOL translation in planetary exploration, several limitations should be noted.

First, the dataset is small in scale, consisting of 11 mission segments derived from 5 unique sources. This work should therefore be viewed as a pilot benchmark intended to motivate further data collection and expansion. Second, the annotations are manually constructed, which introduces potential subjectivity in how natural language descriptions are mapped to logical representations. Third, the evaluation presented in this work is preliminary and qualitative in nature. Future work should explore quantitative evaluation methods, including structural similarity, logical equivalence, and execution-based validation. Finally, the experimental setup relies on direct, single-pass generation with a fixed output length constraint. This setting does not reflect more advanced prompting strategies, decomposition methods, or constrained decoding approaches that may significantly improve performance. 

Despite these limitations, we believe this benchmark provides a useful starting point for studying the intersection of natural language understanding and formal reasoning in long-horizon, structured environments.

\section{CONCLUSION}
We present a pilot benchmark for translating natural language mission descriptions into First-Order Logic, constructed from real-world planetary mission documentation. The dataset captures the complexity of mission phases, temporal dependencies, and operational constraints, and augments these representations with structured predicate vocabularies and typed constants. By focusing on the problem of translating high-level mission knowledge into formal representations, this work highlights an important direction for enabling interpretable and verifiable autonomy. We hope this pilot benchmark will serve as a foundation for future research at the intersection of language understanding and formal reasoning in complex, real-world environments.

% \appendix

% \section{Full pilot benchmark Data}

\end{document}